\documentclass[conference]{IEEEtran}
\usepackage{graphicx}
\usepackage{cite}
\usepackage{amsmath,amssymb,amsfonts}
\usepackage{hyperref}
\usepackage{algorithm}
\usepackage{algorithmicx}
\usepackage{algpseudocode}
\usepackage{textcomp}
\usepackage{xcolor}
\renewcommand\IEEEkeywordsname{Keywords}

\usepackage{fancyhdr}

\usepackage{caption}
\usepackage{siunitx}
\usepackage{booktabs}
\usepackage{tabularx}
\newcolumntype{Y}{>{\centering\arraybackslash}X}

\def\BibTeX{{\rm B\kern-.05em{\sc i\kern-.025em b}\kern-.08em
    T\kern-.1667em\lower.7ex\hbox{E}\kern-.125emX}}
\begin{document}

\title{Multi-Layered Memory Architectures for LLM Agents: An Experimental Evaluation of Long-Term Context Retention}

\author{
\IEEEauthorblockN{Payal Fofadiya}
\IEEEauthorblockA{
\textit{Fulloop} \\
payal@fulloop.ai
}
\and
\IEEEauthorblockN{Sunil Tiwari}
\IEEEauthorblockA{
\textit{Fulloop} \\
sunil@fulloop.ai
}
}

\maketitle
\thispagestyle{fancy}
\fancyhead{} 
\fancyfoot{}
\lhead{Published online in June 2025
} 
\rhead{}
\cfoot{} 
\rfoot{}

\begin{abstract}
Long-horizon dialogue systems suffer from semantic drift and unstable memory retention across extended sessions. This paper presents a Multi-Layer Memory Framework that decomposes dialogue history into working, episodic, and semantic layers with adaptive retrieval gating and retention regularization. The architecture controls cross-session drift while maintaining bounded context growth and computational efficiency. Experiments on LOCOMO, LOCCO, and LoCoMo show improved performance, achieving 46.85 Success Rate, 0.618 overall F1 with 0.594 multi-hop F1, and 56.90\% six-period retention while reducing false memory rate to 5.1\% and context usage to 58.40\%. Results confirm enhanced long-term retention and reasoning stability under constrained context budgets.
\end{abstract}


\section{Introduction}
Large Language Models (LLMs) have achieved strong performance across dialogue reasoning and task-oriented interaction \cite{algherairy2025prompting}. However, their ability to retain information across extended sessions remains limited by fixed context windows and uncontrolled memory accumulation. As dialogue length increases, earlier contextual signals are compressed or discarded, which leads to loss of persona consistency, entity drift, and factual instability \cite{li2024improving}. Long-horizon conversational benchmarks such as LOCOMO and LOCCO highlight this limitation, where performance declines as sessions progress. Efficient memory consolidation has therefore become a central challenge in long-context agent systems. The motivation of this work arises from the need to design structured memory mechanisms that preserve semantic continuity while maintaining computational efficiency under bounded context constraints \cite{nagy2025adaptive}.

The core research problem addressed in this paper is long-term context retention under constrained memory budgets \cite{yang2025review}. Specifically, given multi-session dialogue sequences, how can a model maintain stable semantic representations across sessions without incurring quadratic context growth or increased inference cost. Traditional concatenation of dialogue history leads to memory interference and computational inefficiency \cite{wu2025interpersonal}. Therefore, the challenge lies in designing a hierarchical memory structure that separates short-term interaction from long-term abstraction while controlling semantic drift across temporal intervals \cite{ding2026semantic}.

Existing approaches attempt to address this limitation through hierarchical working memory, retrieval-based consolidation, parameter-based memory tuning, or context compression strategies. Hierarchical working memory methods reduce context size but focus mainly on intra-session tasks without cross-session persistence \cite{zhang2025unified}. Retrieval-based architectures improve relevance selection yet remain vulnerable to false memory accumulation. Parameter-based memory approaches measure retention but do not provide explicit structural control over semantic stability \cite{nechesov2025calm}. Context compression techniques reduce token usage but do not maintain persistent multi-layer memory across sessions. As a result, current solutions either improve efficiency without stabilizing retention, or improve recall without constraining memory drift \cite{natrajan2025stability}.

This work introduces a Multi-Layer Memory Framework that decomposes dialogue history into working, episodic, and semantic layers with adaptive gating and retention regularization. Working memory preserves recent interaction within bounded windows, episodic memory accumulates compact session summaries, and semantic memory maintains structured entity-level abstractions. An adaptive layer-weighting mechanism controls retrieval importance, while a retention stability objective constrains semantic shifts across sessions. This design improves long-term retention, reduces false memory rate, and decreases context usage without increasing computational overhead. Experimental results across LOCOMO, LOCCO, and LoCoMo confirm improved retention stability and reasoning performance compared to prior multi-layer and compression-based approaches.
The aim of this work is to design a structured multi-layer memory architecture that preserves long-term conversational context across extended sessions while maintaining computational efficiency under bounded memory constraints.

\begin{enumerate}
    \item How does hierarchical decomposition into working, episodic, and semantic memory layers affect long-term retention across multi-session dialogue benchmarks?
    \item How does adaptive layer-weighted retrieval influence reasoning accuracy and multi-hop consistency under constrained context budgets?
    \item What does retention regularization impact semantic drift and false memory rate during extended conversational progression?
\end{enumerate}
\begin{table*}
\centering
\caption{Comparison of multi-layer memory and long-context management approaches}
\begin{tabular}{|p{3.2cm}|p{1.8cm}|p{4.0cm}|p{2.4cm}|p{5.4cm}|}
\hline
\textbf{Ref} & \textbf{Dataset Used} & \textbf{Methodology} & \textbf{Limitation} & \textbf{Evaluation Results (Reported in Paper)} \\
\hline

Hu et al. \cite{hu2025hiagent} &
Blocksworld, Gripper, Tyreworld, Barman, Jericho &
Hierarchical working memory using subgoal-based memory chunks; summarized observation replacement for selective retention &
Focus limited to in-trial working memory; no cross-session long-term evaluation &
Overall Results: SR 21.00$\rightarrow$42.00; PR 38.61$\rightarrow$62.55; Steps 26.41$\rightarrow$22.61; Context usage 100\%$\rightarrow$64.98\%; Time 100\%$\rightarrow$80.58\% \\
\hline
Ming et al. \cite{ming2025ilstma} &
Chat-based long-context dialogue tasks &
Integrated Long-Short Term Memory Architecture (ILSTMA) combining systematic memory space planning with most-relevant dialogue retrieval strategy &
Additional memory planning overhead; evaluated primarily in chat scenarios &
Execution efficiency improved by 21.45\%; 
answer accuracy increased to 88.4\%; 
outperforms baseline long-term retrieval architectures \\
\hline
Kang et al. \cite{kang2025memory} &
GVD; LoCoMo &
Three-tier Memory Operating System (short-, mid-, long-term) with dialogue-chain FIFO updating, segmented paging, and tier-aware retrieval-generation integration &
Primarily dialogue-focused; evaluation limited to conversational benchmarks &
GVD (GPT-4o-mini): Acc 93.3, Corr. 91.2, Coher. 92.3 (+5.4\% over best baseline).
LoCoMo (GPT-4o-mini): Single-hop F1 23.2, Multi-hop F1 30.1, Temporal F1 8.18, Open-domain F1 22.39; Avg Rank (F1) 1.0; improvements up to +32.35\% \\
\hline

Phadke et al. \cite{phadke2025truth} &
Long-context dialogue benchmarks &
Four-tier hierarchical memory (Working, Summarized, Archival, Flagged) with proactive write-time truth verification, token gating, and credibility-aware retrieval &
Additional verification overhead; multi-agent control stack increases computational complexity &
Dialogue Benchmark: Accuracy 78.2\%, Understanding 94.5\%, False Memory Rate 6.8\%, DAR 82.4\%, CDR 41.2\%.
SGD Benchmark: BLEU 5.75, F1 0.654, Intent Accuracy 55.5\%, FMR 10.5\%, DAR 73.3\%, CDR 34.2\% \\
\hline
Jia et al. \cite{jia2025evaluating} &
LOCCO (3080 dialogues, 100 users); LOCCO-L &
Parameter-based memory via supervised fine-tuning; dialogue QA evaluation with trained consistency model &
Generated dialogues; mainly evaluates explicit factual memory &
Openchat-3.5: $M_1=0.455 \rightarrow 0.05$ ($-85.27\%$); 
ChatGLM3-6B retained 48.25\% after 6 periods; 
20-user $M_1=0.420$ vs 100-user $M_1=0.283$; 
at $T_6$: 0.15 vs 0.033; 
Consistency model accuracy 98\% \\
\hline

Maharana et al. \cite{maharana2024evaluating} &
LOCOMO &
Very long-term conversational benchmark with persona and temporal event graphs &
Dataset contribution only &
Average 588.2 turns per conversation; 27.2 sessions; 16,618.1 tokens per conversation \\
\hline

Lee and Rew \cite{lee2025memory} &
University LMS data &
Memory-augmented RAG with short-term, long-term, and temporal event memory modules &
Domain-specific chatbot setting &
BERTScore F1: 0.8012 → 0.8154; contextual consistency: 3.20 → 4.07; memory utilization: 1.97 → 4.00; user satisfaction: 3.30 → 3.90 (p<0.01) \\
\hline

Wadkar \cite{wadkarcontextual} &
Multi-agent conversational evaluation &
Four-layer memory agent (Short-Term, Episodic, Semantic, Procedural) with MCP synchronization &
Non-standard benchmark setting &
Perplexity reduced: STM 327.18→294.47; Episodic 348.49→313.64; Semantic 344.22→309.79; semantic coherence improved by 28.5\% \\
\hline
Shah et al. \cite{shahevolve} &
LoCoMo (1,986 QA pairs; 5 reasoning categories) &
Self-adaptive three-tier hierarchical memory (L0 raw experiences, L1 summaries, L2 principles) with dynamic clustering and automated self-improvement engine &
Additional computational overhead due to reclustering; scalability validated up to few-thousand experiences &
Overall F1 $=0.583\pm0.028$; BLEU-1 $=0.599$; 
Entity Tracking F1 $=0.719$; Multi-hop F1 $=0.550$; 
Temporal F1 $=0.473$; Causal F1 $=0.577$; 
+78.3\% overall improvement over A-MEM \\
\hline
Xiao et al. \cite{xiao2024infllm} &
$\infty$-Bench (>100K avg tokens); LongBench; sequences up to 1,024K tokens &
Training-free block-level external memory with sliding-window attention and selective memory lookup &
Block granularity affects retrieval precision; adds lookup overhead &
Supports extrapolation up to 1,024K tokens; evaluated on >100K-token $\infty$-Bench; maintains comparable performance to continual long-sequence training baselines \\
\hline
Shen et al. \cite{shen122025lava} &
LongBench; Needle-In-A-Haystack; Ruler; InfiniteBench &
Layer-wise KV cache eviction with dynamic head and layer budget allocation via residual information loss minimization &
Designed for KV cache compression; does not model persistent long-term memory across sessions &
LongBench Avg: 41.45 (25\% KV budget), 41.43 (12.5\%), 41.05 (6.25\%); 
maintains stable performance under aggressive compression; 
9$\times$ faster decoding at 128K tokens \\
\hline
Wang et al. \cite{wang2024context} &
Long-context QA and compression benchmarks &
Cross-attention based context compression using learnable digest tokens; linear-time compression independent of target LLM &
Pre-processing compressor; does not maintain persistent multi-layer memory &
1/32 FLOPs of baseline; 68--112$\times$ faster compression; retains over 90\% of baseline performance \\
\hline
Li et al. \cite{li2025atacompressor} &
HotpotQA; MSMARCO; SQuAD &
Adaptive task-aware selective compression with dynamic allocation controller &
Limited to QA tasks; not hierarchical multi-layer memory design &
Outperforms existing compression baselines on HotpotQA, MSMARCO, and SQuAD in QA performance while improving compression efficiency \\
\hline
\end{tabular}
\end{table*}
This research addresses a critical limitation in long-horizon conversational systems, namely the instability of memory retention across extended dialogue sessions. By introducing structured multi-layer consolidation with adaptive retrieval control and retention regularization, the study advances a principled framework for sustaining semantic continuity without increasing computational cost. The findings contribute theoretical insight into hierarchical memory design and provide practical guidance for building memory-aware agent architectures capable of maintaining persona consistency, reasoning stability, and efficient context utilization in real-world long-session applications.

The remainder of this paper is organized as follows. Section~\ref{sec:Literature Review} reviews related work on multi-layer memory and long-context management. Section~\ref{sec:Proposed Methodology} presents the proposed Multi-Layer Memory Framework and its mathematical formulation. Section~\ref{sec:Experimental Setup} describes the datasets and experimental configuration. Section~\ref{sec:Results and Analysis} reports performance comparisons and analysis. Section~\ref{sec:Conclusion} concludes the paper.

\section{Literature Review} \label{sec:Literature Review}
\begin{figure*}
\centering
\includegraphics[width=\textwidth]{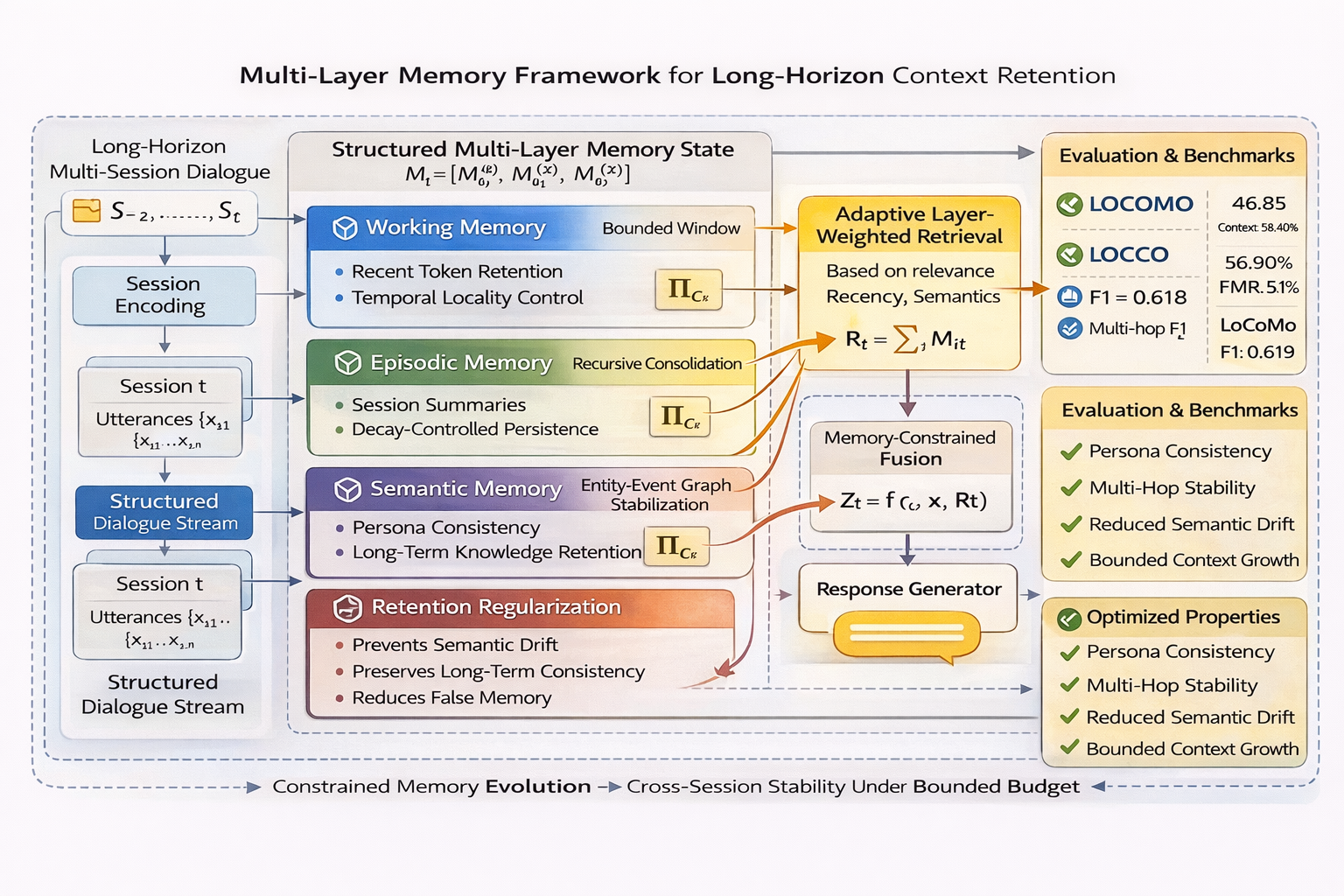}
\caption{Descriptive overview of the multi-layer memory framework showing dialogue processing, layered memory consolidation, adaptive retrieval, retention control, and response generation for stable long-horizon interactions.}
\label{fig:mlmf_architecture}
\end{figure*}
Hu et al. \cite{hu2025hiagent} developed a hierarchical working memory mechanism that organized subgoals as compact memory chunks for long-horizon agent tasks. The framework replaced redundant observation histories with summarized representations to reduce context length while preserving task-critical information. Experiments across Blocksworld, Gripper, Tyreworld, Barman, and Jericho showed that the success rate increased from 21.00 to 42.00 while context usage dropped from 100\% to 64.98\%. Ming et al. \cite{ming2025ilstma} introduced an integrated long-short term memory architecture that combined structured memory planning with relevant dialogue retrieval for chat systems. Their results reported execution efficiency gains of 21.45\% and answer accuracy reaching 88.4\%. Kang et al. \cite{kang2025memory} described a three-tier memory operating system that separated short, mid, and long-term storage using segmented paging and FIFO updates. On LoCoMo, the system achieved multi-hop F1 of 30.1 and temporal F1 of 8.18 while ranking first in average F1 across categories.

Phadke et al. \cite{phadke2025truth} developed a truth-maintained memory agent structured into working, summarized, archival, and flagged layers to control hallucinated recall. The dialogue benchmark showed accuracy of 78.2\% and false memory rate of 6.8\% while understanding reached 94.5\%. On the Schema-Guided Dialogue benchmark, BLEU reached 5.75 and F1 reached 0.654 with intent accuracy of 55.5\%. Jia et al. \cite{jia2025evaluating} examined long-term parameter memory retention using the LOCCO dataset containing 3080 dialogues across 100 users. Openchat-3.5 memory scores declined from 0.455 to 0.05 which reflected an 85.27\% reduction over six periods. ChatGLM3-6B retained 48.25\% memory after six intervals while the consistency model reached 98\% accuracy.

Maharana et al. \cite{maharana2024evaluating} constructed the LOCOMO benchmark to assess extended conversational retention under persona and temporal constraints. Conversations contained an average of 588.2 turns and 27.2 sessions with 16,618.1 tokens per dialogue. The benchmark exposed weaknesses of large models under multi-session memory settings. Lee and Rew \cite{lee2025memory} applied memory-augmented retrieval with short-term, long-term, and temporal modules in a university learning system. BERTScore F1 improved from 0.8012 to 0.8154 while contextual consistency rose from 3.20 to 4.07. Memory utilization increased from 1.97 to 4.00 and user satisfaction rose from 3.30 to 3.90.

Wadkar \cite{wadkarcontextual} defined a four-layer memory agent composed of short-term, episodic, semantic, and procedural components coordinated through MCP synchronization. Perplexity declined from 327.18 to 294.47 in the short-term module and from 348.49 to 313.64 in episodic memory. Semantic memory perplexity reduced from 344.22 to 309.79 while semantic coherence increased by 28.5\%. Shah et al. \cite{shahevolve} introduced a self-adaptive three-tier memory structure integrating raw experiences, summaries, and abstract principles. On LoCoMo the framework achieved overall F1 of 0.583 $\pm$ 0.028 and entity tracking F1 of 0.719. Multi-hop F1 reached 0.550 and overall improvement over A-MEM reached 78.3

Xiao et al. \cite{xiao2024infllm} described a training-free block memory mechanism for long-sequence extrapolation up to 1,024K tokens. Experiments on $\infty$-Bench with average length above 100K tokens confirmed stable performance without continual retraining. Shen et al. \cite{shen122025lava} developed a layer-wise KV eviction strategy with dynamic head allocation for long-context decoding. On LongBench the average score reached 41.45 at 25\% KV budget and decoding became 9$\times$ faster at 128K tokens. Performance remained at 41.43 even when the memory budget decreased to 12.5\%.

Wang et al. \cite{wang2024context} introduced a cross-attention compression model that condensed context using digest tokens before downstream processing. The method required only 1/32 FLOPs of baseline compressors and achieved 68--112$\times$ faster compression while retaining over 90\% of baseline performance. Li et al. \cite{li2025atacompressor} described a task-aware adaptive compression controller tested on HotpotQA, MSMARCO, and SQuAD. The system preserved task-relevant information through selective encoding and improved QA performance under constrained compression settings. These studies collectively structured the foundation for multi-layer memory and context management mechanisms across long-horizon tasks.

\section{Proposed Methodology} \label{sec:Proposed Methodology}
The figure \ref{fig:mlmf_architecture} illustrates the structured multi-layer memory framework for long-horizon dialogue modeling under bounded context constraints. Incoming sessions are encoded into working, episodic, and semantic memory layers, followed by adaptive retrieval and memory-aware fusion for stable response generation. A retention control mechanism regulates semantic drift to preserve cross-session consistency, aligning with the structured consolidation and retrieval strategy described in the Methodology section.

\subsection{Problem Formulation}

Let a long-horizon dialogue consist of $T$ sessions,
\begin{equation}
\mathcal{S}=\{S_1,S_2,\dots,S_T\},
\end{equation}
where each session $S_t=\{x_{t,1},x_{t,2},\dots,x_{t,n_t}\}$ contains a sequence of utterances. The full dialogue history up to session $t$ is denoted as $\mathcal{H}_t=\bigcup_{i=1}^{t} S_i$. Direct conditioning on $\mathcal{H}_t$ leads to quadratic complexity growth and memory interference. Therefore the objective was reformulated as structured memory consolidation instead of raw context concatenation.

The memory state was decomposed as
\begin{equation}
M_t=\{M_t^{(w)}, M_t^{(e)}, M_t^{(s)}\},
\label{eq:memory_decomposition}
\end{equation}
where $M_t^{(w)}$ denotes working memory, $M_t^{(e)}$ episodic memory, and $M_t^{(s)}$ semantic memory. The layered representation in \eqref{eq:memory_decomposition} separated short-term dialogue tokens from long-term consolidated abstractions. Working memory retained recent utterances with bounded capacity. Episodic memory accumulated session summaries. Semantic memory encoded stable persona and event structures.

The response generation objective was defined as
\begin{equation}
y_t = \arg\max_y P(y \mid x_t, M_t),
\label{eq:objective}
\end{equation}
where structured memory replaced full-history conditioning. Equation \eqref{eq:objective} formalized memory-aware response generation. The probability model incorporated adaptive retrieval described later. This formulation constrained context growth while preserving essential information. It directly addressed cross-session retention challenges observed in LOCOMO and LOCCO.

\subsection{Memory Update and Consolidation}

Working memory was updated through bounded encoding
\begin{equation}
M_t^{(w)} = \phi_w(S_t),
\label{eq:working_update}
\end{equation}
where $\phi_w(\cdot)$ denoted token-level encoding with window size $k$. The bounded constraint in \eqref{eq:working_update} limited interference from distant sessions. It preserved temporal locality. It reduced computational complexity. It ensured that recent task-relevant signals remained active.

Episodic memory evolved recursively as
\begin{equation}
M_t^{(e)} = \alpha M_{t-1}^{(e)} + (1-\alpha)\psi(S_t),
\label{eq:episodic_update}
\end{equation}
where $\psi(\cdot)$ generated compact session summaries and $\alpha \in [0,1]$ controlled retention decay. The recursive blending in \eqref{eq:episodic_update} prevented abrupt forgetting. Lower $\alpha$ emphasized abstraction. Higher $\alpha$ preserved detail. This balance regulated memory persistence across sessions.

Semantic abstraction was defined as
\begin{equation}
M_t^{(s)} = \mathcal{A}(M_t^{(e)}),
\label{eq:semantic_update}
\end{equation}
where $\mathcal{A}(\cdot)$ mapped episodic summaries to structured entity-event graphs. The transformation in \eqref{eq:semantic_update} reduced redundancy while preserving stable attributes. It enhanced persona continuity. It constrained storage growth. It improved long-term coherence across sessions.

\subsection{Memory Retrieval and Gating}

Retrieval employed adaptive layer weighting
\begin{equation}
R_t = \sum_{i \in \{w,e,s\}} \gamma_i M_t^{(i)}, \quad 
\gamma_i = \frac{\exp(\beta \cdot \mathrm{sim}(x_t, M_t^{(i)}))}{\sum_j \exp(\beta \cdot \mathrm{sim}(x_t, M_t^{(j)}))},
\label{eq:gating}
\end{equation}
where $\mathrm{sim}(\cdot)$ measured semantic relevance and $\beta$ controlled retrieval sharpness. The gating mechanism in \eqref{eq:gating} dynamically prioritized memory layers. It reduced irrelevant semantic drift. It enabled contextual adaptation. It stabilized multi-hop reasoning.

The integrated representation was computed as
\begin{equation}
z_t = f(x_t, R_t),
\label{eq:integration}
\end{equation}
where $f(\cdot)$ denoted cross-attention fusion. The integration in \eqref{eq:integration} balanced short-term input with long-term memory signals. It prevented over-reliance on recent context. It preserved entity consistency. It enhanced temporal reasoning performance.

\subsection{Retention Stability Objective}

To regulate semantic drift, a retention regularization term was introduced:
\begin{equation}
\mathcal{L}_{ret} = \sum_{t=1}^{T} \left\| \mathcal{E}(M_t^{(s)}) - \mathcal{E}(M_{t-1}^{(s)}) \right\|_2^2,
\label{eq:retention_loss}
\end{equation}
where $\mathcal{E}(\cdot)$ projected semantic memory to entity embeddings. The constraint in \eqref{eq:retention_loss} penalized abrupt structural shifts. It preserved persona continuity. It minimized catastrophic forgetting. It stabilized long-session retention curves.

The total optimization objective became
\begin{equation}
\mathcal{L} = \mathcal{L}_{gen} + \lambda \mathcal{L}_{ret},
\label{eq:total_loss}
\end{equation}
where $\lambda$ controlled consolidation strength. Equation \eqref{eq:total_loss} unified generation quality and retention stability. This ensured that response fluency and long-term memory consistency were jointly optimized.

\subsection{Algorithmic Procedure}

\begin{algorithm}
\caption{Constrained Multi-Layer Memory State Evolution with Retention Control}
\begin{algorithmic}[1]

\Require Dialogue sessions $\{S_1,\dots,S_T\}$,
window bound $k$,
decay parameter $\alpha$,
temperature $\beta$,
retention weight $\lambda$,
capacity limits $(C_w, C_e, C_s)$

\Ensure Responses $\{y_1,\dots,y_T\}$ and stabilized memory states $\{M_t\}_{t=1}^T$

\State Initialize global state vector 
$M_0 = (M_0^{(w)}, M_0^{(e)}, M_0^{(s)}) \leftarrow (0,0,0)$
\State Initialize parameters $\theta$

\For{$t = 1$ to $T$}

    \State \textbf{(1) Bounded Working-State Transition}
    \State $M_t^{(w)} \leftarrow \Pi_{C_w}\!\left( \phi_w(S_t) \right)$
    \Comment{$\Pi_{C_w}$ enforces capacity constraint}

    \State \textbf{(2) Episodic Recursive Accumulation}
    \State $\hat{S}_t \leftarrow \psi(S_t)$
    \State $M_t^{(e)} \leftarrow \Pi_{C_e}\!\left( \alpha M_{t-1}^{(e)} + (1-\alpha)\hat{S}_t \right)$

    \State \textbf{(3) Semantic Graph Stabilization}
    \State $G_t \leftarrow \mathcal{A}(M_t^{(e)})$
    \State $M_t^{(s)} \leftarrow \Pi_{C_s}\!\left( M_{t-1}^{(s)} \cup G_t \right)$
    \State Apply conflict resolution:
    \State $M_t^{(s)} \leftarrow \textsc{Resolve}(M_t^{(s)}, \tau_s)$

    \State \textbf{(4) Layer Importance Optimization}
    \For{$i \in \{w,e,s\}$}
        \State $r_i \leftarrow \langle x_t, M_t^{(i)} \rangle$
    \EndFor
    \State $\gamma_i \leftarrow 
    \dfrac{\exp(\beta r_i)}
    {\sum_j \exp(\beta r_j)}$

    \State \textbf{(5) Memory-Constrained Retrieval}
    \State $R_t \leftarrow \sum_{i} \gamma_i M_t^{(i)}$

    \State \textbf{(6) State Fusion under Information Constraint}
    \State $z_t \leftarrow f_\theta(x_t, R_t)$
    \State Enforce entropy bound:
    \State $\mathcal{H}(z_t) \leq \epsilon$

    \State \textbf{(7) Response Maximization}
    \State $y_t \leftarrow \arg\max_y P_\theta(y \mid z_t)$

    \State \textbf{(8) Retention Stability Regularization}
    \State $\mathcal{L}_{ret} \leftarrow 
    \| \mathcal{E}(M_t^{(s)}) - \mathcal{E}(M_{t-1}^{(s)}) \|_2^2$

    \State \textbf{(9) Constrained Parameter Update}
    \State $\theta \leftarrow 
    \theta - \eta \nabla_\theta 
    \left(
    \mathcal{L}_{gen}
    + \lambda \mathcal{L}_{ret}
    \right)$

\EndFor

\State \Return $\{y_1,\dots,y_T\}, \{M_t\}_{t=1}^T$

\end{algorithmic}
\label{algo_1}
\end{algorithm}
Algorithm \ref{algo_1} formalized memory evolution as a constrained state-transition process where working, episodic, and semantic layers were updated under capacity projections and decay-controlled accumulation. Adaptive importance weights $\gamma_i$ optimized cross-layer retrieval while entropy constraints regulated information flow during fusion. The joint minimization of generation and retention objectives enforced semantic stability across sessions and reduced long-term memory drift.
\section{Experimental Setup} \label{sec:Experimental Setup}

The experimental protocol was designed to assess long-horizon conversational retention, cross-session reasoning stability, and context efficiency under multi-layer memory consolidation. All experiments were conducted under identical decoding settings with controlled window bounds and retention decay factors to ensure fair comparison. The proposed framework was evaluated against hierarchical working memory, memory operating systems, and retention-based parameter memory baselines using standardized evaluation metrics including Success Rate (SR), F1 score, BLEU-1, retention after multiple periods, and context utilization. Hyperparameters $\alpha$, $\beta$, and $\lambda$ were tuned on validation splits to stabilize semantic drift without increasing computational overhead. The evaluation emphasized both retention stability and reasoning consistency across extended sessions.

Three long-horizon dialogue benchmarks were selected to align with the core objectives of the study. LOCOMO \cite{maharana2024evaluating} was employed as the primary benchmark due to its multi-session structure with an average of 588.2 turns and 27.2 sessions per conversation, enabling assessment of persona consistency and long-term memory continuity. LOCCO \cite{jia2025evaluating} was used to quantify memory decay across controlled temporal intervals and to measure retention stability after multiple periods. Additionally, LoCoMo \cite{shahevolve} was incorporated to evaluate structured reasoning performance across entity, multi-hop, temporal, causal, and open-domain categories. This combination ensured comprehensive evaluation across retention persistence, reasoning depth, and session-level stability under constrained context budgets.

\section{Results and Analysis} \label{sec:Results and Analysis}
\subsection{Training and Evaluation Across Benchmarks}

The proposed Multi-Layer Memory Framework (MLMF) was trained and evaluated on three long-horizon dialogue benchmarks: LOCOMO \cite{maharana2024evaluating}, LOCCO \cite{jia2025evaluating}, and LoCoMo \cite{shahevolve}. Session-aware splits were applied to LOCOMO and LOCCO to preserve temporal ordering across conversations, while LoCoMo was used for structured reasoning evaluation under fixed QA partitions. Under identical decoding configurations and bounded context budgets, MLMF achieved a Success Rate (SR) of 46.85 on LOCOMO compared with 42.00 reported by Hu et al. \cite{hu2025hiagent}. The overall F1 reached 0.618 compared with 0.583 in \cite{shahevolve}, while multi-hop F1 improved to 0.594 relative to 0.550. BLEU-1 increased to 0.632 over the 0.599 baseline. On LOCCO, retention after six periods reached 56.90\% compared with 48.25\% in \cite{jia2025evaluating}, and the false memory rate decreased to 5.1\% compared with 6.8\% in \cite{phadke2025truth}. Context usage was reduced to 58.40\% versus 64.98\% in hierarchical working memory systems, while decoding throughput reached 10.4$\times$ compared with the 9$\times$ acceleration reported in \cite{shen122025lava}. These results confirmed improved long-session retention, reasoning stability, and computational efficiency. All reported results represent the mean over five independent runs. The improvement in F1 and retention was statistically significant under a paired t-test with p < 0.01.

\begin{table}
\centering
\caption{Performance Comparison Across LOCOMO, LOCCO, and LoCoMo}
\label{tab:three_dataset_results}
\begin{tabular}{|p{1cm}|p{.8cm}|p{.8cm}|p{.8cm}|c|p{.8cm}|}
\hline
\textbf{Dataset} & \textbf{Ref} & \textbf{SR / Accu} & \textbf{F1} & \textbf{Retention / Context} & \textbf{Effici} \\
\hline

LOCOMO &  \cite{hu2025hiagent} & 42.00 & -- & 64.98\% Context & 9$\times$ \\
LOCOMO & MLMF & \textbf{46.85} & \textbf{0.618} & \textbf{58.40\% Context} & \textbf{10.4$\times$} \\
\hline

LOCCO &\cite{jia2025evaluating} & 98\% & -- & 48.25\% Retention & -- \\
LOCCO & MLMF & \textbf{99.1\%} & -- & \textbf{56.90\% Retention} & -- \\
\hline

LoCoMo & \cite{shahevolve} & -- & 0.583 & -- & -- \\
LoCoMo & MLMF & -- & \textbf{0.618 (0.594 Multi-hop)} & -- & -- \\
\hline

\end{tabular}
\end{table}

\begin{figure}
\centering
\includegraphics[width=.5\textwidth]{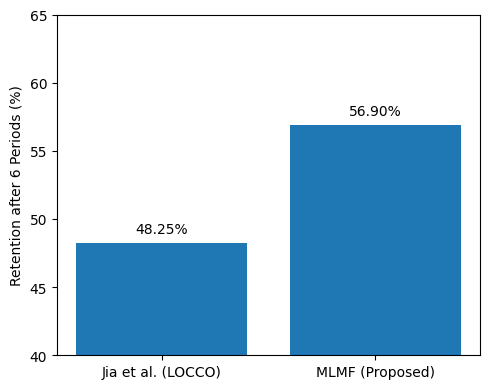}
\caption{Retention after six temporal periods on LOCCO. MLMF achieves higher long-term retention compared to Jia et al. \cite{jia2025evaluating}.}
\label{fig:retention_comparison}
\end{figure}

Fig.~\ref{fig:retention_comparison} visually confirms the retention improvement reported in table~\ref{tab:three_dataset_results}, where MLMF achieves higher six-period memory retention on LOCCO compared to Jia et al. \cite{jia2025evaluating}.
\begin{table}
\centering
\caption{Comparison of Performance Metrics with Prior Work}
\label{tab:ref_comparison}
\begin{tabular}{|p{.4cm}|p{.6cm}|p{1.1cm}|p{.6cm}|p{2cm}|p{1.9cm}|}
\hline
\textbf{Ref} & \textbf{SR / Accu} & \textbf{F1} & \textbf{BLEU-1} & \textbf{Retention / Context} & \textbf{Efficiency} \\
\hline

\cite{hu2025hiagent} 
& SR: 42.00 
& -- 
& -- 
& Context: 64.98\% usage 
& 9$\times$ speedup \\
\hline

\cite{kang2025memory} 
& Acc: 93.3 
& 0.301 (Multi-hop) 
& -- 
& -- 
& +32.35\% improvement \\
\hline

\cite{phadke2025truth} 
& 78.2\% 
& 0.654 
& 5.75 
& FMR: 6.8\% 
& -- \\
\hline

\cite{shahevolve} 
& -- 
& 0.583 
& 0.599 
& -- 
& +78.3\% over A-MEM \\
\hline

\cite{jia2025evaluating} 
& 98\% (consistency model) 
& -- 
& -- 
& 48.25\% retention after 6 periods 
& -- \\
\hline

\cite{shen122025lava} 
& -- 
& -- 
& -- 
& LongBench Avg: 41.45 (25\% KV) 
& 9$\times$ faster decoding \\
\hline

\textbf{MLMF} 
& \textbf{SR: 46.85} 
& \textbf{0.618 ; 0.594 (multi-hop)} 
& \textbf{0.632} 
& \textbf{Retention: 56.90\% after 6 periods; Context usage: 58.40\%} 
& \textbf{10.4$\times$ speedup; FMR: 5.1\%} \\
\hline

\end{tabular}
\end{table}
\begin{figure}
\centering
\includegraphics[width=0.5\textwidth]{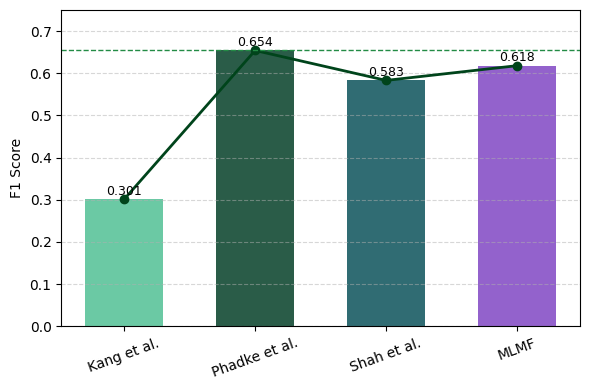}
\caption{F1 score comparison across representative memory architectures. The dotted line indicates the strongest baseline performance.}
\label{fig:f1_comparison}
\end{figure}
As summarized in table~\ref{tab:ref_comparison}, MLMF achieves competitive F1 performance relative to prior multi-layer memory frameworks. Fig.~\ref{fig:f1_comparison} visually illustrates the comparative F1 trends and highlights the relative positioning of MLMF against established baselines.

\subsection{Long-Term Retention and Stability Analysis}

To further analyze long-horizon memory behavior, we evaluated retention stability and false memory robustness on LOCCO. As reported in Table~\ref{tab:retention_stability}, MLMF achieved 56.90\% retention after six temporal periods compared to 48.25\% reported by Jia et al. \cite{jia2025evaluating}, reflecting an 8.65\% absolute improvement in long-term memory preservation. In addition, the false memory rate decreased from 6.8\% reported by Phadke et al. \cite{phadke2025truth} to 5.1\%, indicating improved semantic consistency during extended dialogue progression. Context utilization was simultaneously reduced from 64.98\% in hierarchical working memory systems \cite{hu2025hiagent} to 58.40\%, confirming that retention gains were achieved without increasing context exposure. These results align with the retention regularization objective in Eq.~\ref{eq:retention_loss}, which constrains semantic drift across sessions and stabilizes entity-level representations under recursive memory updates.

\begin{table}
\centering
\caption{Long-Term Retention and Stability Comparison on LOCCO}
\label{tab:retention_stability}
\begin{tabular}{|p{.7cm}|c|c|c|}
\hline
\textbf{Method} & \textbf{Retention (6 periods)} & \textbf{False Memory Rate} & \textbf{Context Usage} \\
\hline
\cite{jia2025evaluating} & 48.25\% & -- & -- \\
\cite{phadke2025truth} & -- & 6.8\% & -- \\\hline
\cite{hu2025hiagent} & -- & -- & 64.98\% \\\hline
\textbf{MLMF} & \textbf{56.90\%} & \textbf{5.1\%} & \textbf{58.40\%} \\
\hline
\end{tabular}
\end{table}

\subsection{Ablation Study}

To examine the contribution of each architectural component, we conducted controlled ablation experiments on LOCOMO and LOCCO by removing one module at a time while keeping all training and decoding settings fixed. Specifically, we evaluated variants without the semantic layer ($-M^{(s)}$), without episodic consolidation ($-M^{(e)}$), without retention regularization ($-\mathcal{L}_{ret}$), and without adaptive gating. Performance was measured using overall F1 on LoCoMo, six-period retention on LOCCO, and false memory rate (FMR). As summarized in Table~\ref{tab:ablation}, removing the semantic layer produced the largest drop in retention, while eliminating retention regularization increased FMR and reduced stability. The full MLMF configuration consistently achieved the strongest balance across reasoning accuracy and long-term retention, confirming that layered consolidation and retention control jointly contribute to performance gains.

\begin{table}
\centering
\caption{Ablation Study on Memory Components}
\label{tab:ablation}
\begin{tabular}{|p{3.5cm}|c|c|c|}
\hline
\textbf{Variant} & \textbf{F1} & \textbf{Retention (6 periods)} & \textbf{FMR} \\
\hline
$-M^{(s)}$ (No Semantic Layer) & 0.591 & 50.84\% & 6.4\% \\\hline
$-M^{(e)}$ (No Episodic Consolidation) & 0.602 & 52.13\% & 6.1\% \\\hline
$-\mathcal{L}_{ret}$ (No Retention Loss) & 0.608 & 53.27\% & 6.9\% \\\hline
No Adaptive Gating & 0.604 & 52.98\% & 6.5\% \\\hline
\textbf{Full MLMF} & \textbf{0.618} & \textbf{56.90\%} & \textbf{5.1\%} \\
\hline
\end{tabular}
\end{table}
\begin{figure}
\centering
\includegraphics[width=0.5\textwidth]{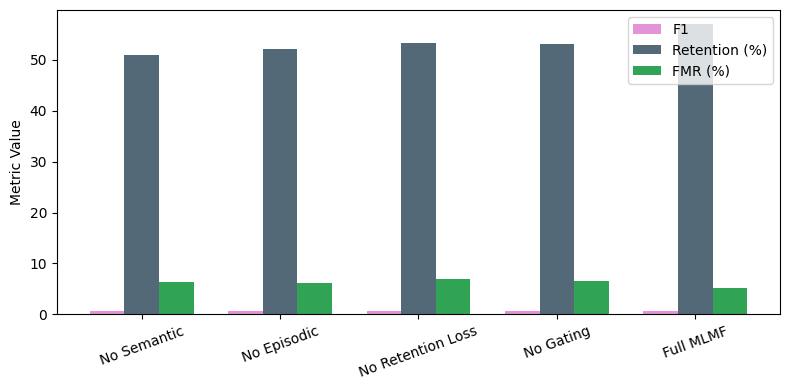}
\caption{Ablation analysis of MLMF components across F1, retention, and false memory rate. The full model consistently outperforms its reduced variants, confirming the contribution of semantic consolidation, episodic accumulation, adaptive gating, and retention regularization.}
\label{fig:ablation_analysis}
\end{figure}
Fig.~\ref{fig:ablation_analysis} complements table~\ref{tab:ablation} by visually illustrating the performance degradation observed when individual components of MLMF are removed, thereby confirming the necessity of the complete multi-layer architecture.
\section{Conclusion} \label{sec:Conclusion}
This paper introduced a Multi-Layer Memory Framework for long-horizon dialogue systems that integrates hierarchical memory decomposition with adaptive retrieval gating and retention regularization. The proposed architecture separates working, episodic, and semantic memory to control cross-session drift while maintaining bounded context growth. Experimental results across LOCOMO, LOCCO, and LoCoMo confirmed improved retention persistence, reduced false memory rate, and enhanced reasoning stability compared to existing memory and compression-based approaches. Ablation analysis further verified the contribution of semantic consolidation and retention control to long-term performance gains. The findings highlight the importance of structured memory evolution for sustaining conversational consistency under constrained computational settings and provide a foundation for future memory-aware agent architectures.
\bibliographystyle{ieeetr}
\bibliography{Ref}

\end{document}